\documentclass[journal,twoside,web]{ieeecolor}
\usepackage{tmi}
\usepackage{cite}
\usepackage{amsmath,amssymb,amsfonts}
\usepackage{algorithmic}
\usepackage{graphicx}
\usepackage{textcomp}
\usepackage{booktabs}
\usepackage{multirow}
\def\BibTeX{{\rm B\kern-.05em{\sc i\kern-.025em b}\kern-.08em
    T\kern-.1667em\lower.7ex\hbox{E}\kern-.125emX}}
\begin{document}
\title{Detecting, Localising and Classifying Polyps from Colonoscopy Videos using Deep Learning}
\author{Yu Tian,  Leonardo Zorron Cheng Tao Pu, Yuyuan Liu, Gabriel Maicas, Johan W. Verjans, Alastair D. Burt, Seon Ho Shin, Rajvinder Singh, Gustavo Carneiro
\thanks{This work was partially supported by the Australian Research Council's Discovery Projects funding scheme (project DP180103232). Gustavo Carneiro is the recipient of an Australian Research Council Future Fellowship (FT190100525).}
\thanks{Yu Tian, Gustavo Carneiro, Yuyuan Liu, and Gabriel Maicas are with the Australian Institute for Machine Learning, University of Adelaide,  Adelaide, Australia. Yu Tian is the corresponding author with email address: yu.tian01@adelaide.edu.au}
\thanks{Yu Tian and Johan W. Verjans are with the South Australian Health and Medical Research Institute, Adelaide SA 5000, Australia.}
\thanks{Leonardo Zorron Cheng Tao Pu is with the Department of Gastroenterology at Austin Health, Heidelberg, Victoria, Australia.}
\thanks{Alastair D. Burt is with the Faculty of Health and Medical Sciences (UoA), Translational and Clinical Research Institute (NU), University of Adelaide, Newcastle University, Adelaide, Australia (UoA); Newcastle, UK (NU).}
\thanks{Seon Ho Shin and Rajvinder Singh are with the Lyell McEwin Hospital, University of Adelaide, Elizabeth Vale, Adelaide SA, Australia.}}

\maketitle

\begin{abstract}
In this paper, we propose and analyse a system that can automatically detect, localise and classify polyps from colonoscopy videos.  
The detection of frames with polyps is formulated as a few-shot anomaly classification problem, where 
the training set is highly imbalanced with the large majority of frames consisting of normal images and a small minority comprising frames with polyps.
Colonoscopy videos may contain blurry images and frames displaying feces and water jet sprays to clean the colon -- such frames can mistakenly be detected as anomalies, so we have implemented a classifier to reject these two types of frames before polyp detection takes place.
Next, given a frame containing a polyp, our method localises (with a bounding box around the polyp) and classifies it into five different classes.
Furthermore, we study a method to improve the reliability and interpretability of the classification result using uncertainty estimation and classification calibration. 
Classification uncertainty and calibration not only help improve classification accuracy by rejecting low-confidence and high-uncertain results, but can be used by doctors to decide how to decide on the classification of a polyp.
All the proposed detection, localisation and classification methods are tested using large data sets and compared with relevant baseline approaches.
\end{abstract}

\begin{IEEEkeywords}
Deep learning, Colonoscopy, Anomaly Detection, Polyp Detection, Polyp Classification, Classification Uncertainty, Classification Calibration
\end{IEEEkeywords}

\section{Introduction}
\label{sec:introduction}
\IEEEPARstart{C}{olorectal} cancer (CRC) is the third most commonly diagnosed cancer in the world and recent studies show that CRC will increase by 60\% by 2030 to more than 2.2 million new cases and 1.1 million cancer deaths~\cite{arnold2017global}. 
These numbers were used as an incentive for the introduction of CRC screening programs in several countries, such as in Australia~\footnote{Bowel Cancer Australia. A Colonoscopy Wait-time and Performance Guarantee. Available at: https://www.bowelcanceraustralia.org/earlydetection/a-colonoscopy-wait-time-and-performance-guarantee.}. An important step in all CRC screening programs is colonoscopy, where the goal is to detect and classify malignant or pre-malignant polyps using a camera that is inserted into the large bowel.
Accurate early polyp detection and classification may improve survival rates~\cite{siegel2014colorectal}, but such accuracy is affected by several human factors, such as expertise and fatigue~\cite{pu2020computer,van2006polyp,pu2019prospective}. 
Hence, a system that can accurately detect and classify polyps during a colonoscopy exam has the potential to improve colonoscopists' efficacy~\cite{tian2020few,liu2020photoshopping,pu2020computer} (according to the American Society for Gastrointestinal Endoscopy, an accurate classification represents an agreement of over 90\% with pathologic assessment).
Furthermore, polyp classification during colonoscopy can be advantageous since it can~\cite{pu2020computer}: 1) reduce the need for histopathologic assessment after resecting pre-malignant polyps with a small risk of an invasive component, 2) avoid resection of benign polyps, 3) determine the most appropriate resection method, 4) enable the estimation of a follow-up period by the end of the procedure, and 5) reduce complications associated with unnecessary polypectomies.

\begin{figure}[t]
\begin{center}
\includegraphics[width=0.49\textwidth]{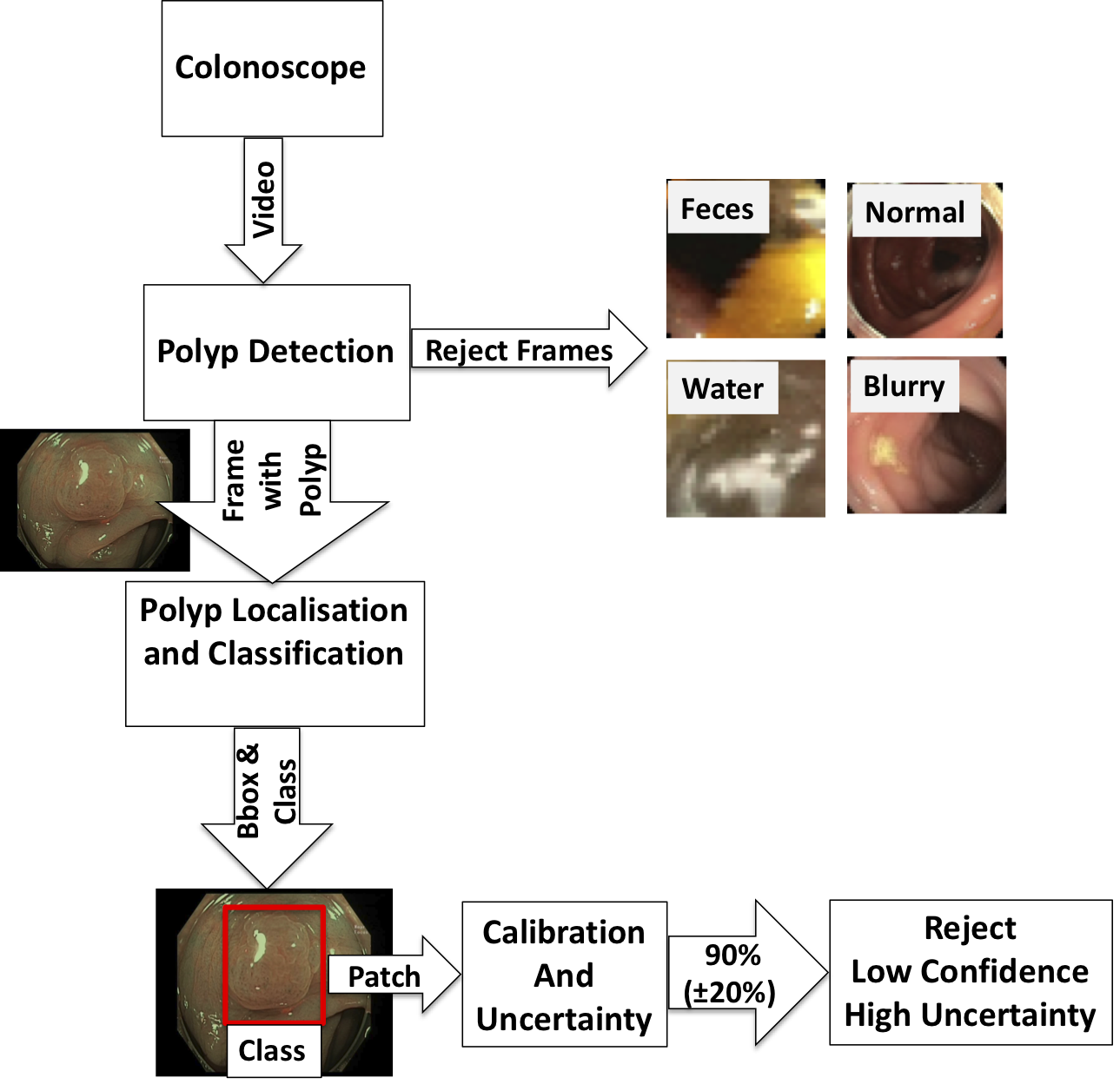} 
\end{center}
\caption{Detection, localisation and classification (with confidence calibration and uncertainty estimation) of polyps from colonoscopy videos.  In the polyp detection phase, frames that show normal tissue, blurry images, feces and water jet sprays are rejected. Then, we localise and classify the polyps using classification uncertainty and calibration.  The last step consists of rejecting classifications with low confidence and high uncertainty to improve classification accuracy and model interpretability.}
\label{fig:motivation}
\end{figure}

As shown in Fig.~\ref{fig:motivation}, automated polyp classification depends on polyp detection, localisation and classification.
Polyp detection consists of the identification of colonoscopy video frames containing polyps, and the rejection of frames containing normal tissue, blurry frames, and frames showing feces or water jet sprays to clean the colon.  The rejection of blurry frames is reached with image processing tools and the rejection of frames showing feces or water jet sprays is achieved with a binary classifier.
The classification of frames containing normal tissue or polyps is formulated 
as a few-shot anomaly detection~\cite{liu2020photoshopping,tian2020few} using a training set that contains a large percentage of normal (i.e., healthy) and a very small proportion abnormal (i.e., unhealthy with polyps) samples.  
Our formulation is designed to detect unforeseen types of polyps not present in the training.  
For instance, if the training images contain only a sub-set of all possible polyp types, the detection should be able to identify all types of polyps, even though they were not included in the training set.
Next, given a frame containing a polyp,  the localisation and classification of the polyp in the image is performed simultaneously~\cite{tian2019oneshot} using state-of-the-art deep learning models~\cite{litjens2017survey}, which are typically accurate, but poorly calibrated~\cite{guo2017calibration} 
and without an uncertainty estimation~\cite{gal2016dropout,gal2017concrete,kendall2017uncertainties}.  
These two issues must be addressed to increase the reliability of the system~\cite{carneiro2020deep}.

In this paper, we integrate recently proposed few-shot anomaly detection methods designed to perform the detection of frames containing polyps from colonoscopy videos~\cite{tian2020few,liu2020photoshopping} with a method that rejects frames containing blurry images, feces and water jet sprays.  
Taking images containing polyps, we analyse the results of our recently proposed method that simultaneously localises and classifies polyps into five classes~\cite{tian2019oneshot}.  
Finally, we analyse the results of our recently proposed polyp classifier~\cite{carneiro2020deep} that outputs classification uncertainty~\cite{settles2012active,gal2016dropout,gal2017concrete,kendall2017uncertainties} and calibrated confidence~\cite{guo2017calibration}.
We also demonstrate that classification accuracy increases by rejecting low-confidence and uncertain classifications.
For each of the proposed methods, we present thorough experiments in multiple data sets to provide evidence of their functionality.

\subsection{Literature Review}
\label{sec:lit_review}


\subsubsection{Polyp Detection}

The detection of polyps is typically addressed using binary classifiers~\cite{korbar2017deep} trained with a large training set containing images without polyps (i.e. normal) and with polyps (i.e. abnormal). 
In this context, it is natural to explore imbalanced learning algorithms~\cite{li2019overfitting,lin2018focal,ren2018learning} because the number of normal images tends to overwhelm that of the abnormal ones, but the extreme imbalanced proportion of normal and abnormal images tend to harm the ability of these algorithms to handle imbalanced learning (e.g., colonoscopy data sets can have 3 to 4 orders of magnitude more normal than abnormal images~\cite{tian2020few}).  
Furthermore, colonoscopy frames may contain many distractors, such as blurry images, water jet sprays and feces~\cite{liu2020photoshopping}. If not present in the training set of the normal class, such images can be mistakenly classified as abnormal. 

An alternative approach to binary classification is based on zero-shot anomaly detection trained using normal images, exclusively~\cite{gong2019memorizing,makhzani2015adversarial,zong2018deep,schlegl2019f,schlegl2017unsupervised,perera2019ocgan,liu2020photoshopping}), where the idea is to learn the distribution of normal images, and abnormal images are detected based on how they fail to fit in this distribution.
Zero-shot anomaly detection works well when what characterises an anomaly covers a relatively large area of the image, which is not always the case for polyps.  
Therefore, using a few abnormal images (less than $100$) to train anomaly detection methods enables the implementation of better polyp detectors~\cite{tian2020few}, referred to as few-shot anomaly detection.

Few-shot anomaly detection has been proposed for non-medical image analysis problems~\cite{pang2018learning}, where their main challenge is on how to deal with the high-dimensionality of the image space.  
Our method~\cite{tian2020few} solves this high-dimensional issue by relying on deep infomax (DIM)~\cite{hjelm2018learning} to reduce the dimensionality of the feature space.

\subsubsection{Polyp Localisation and Classification}

Once frames containing polyps have been identified, the next stage is to localise and classify the polyp. The most straightforward approach consists of a two-stage process that first localises the polyp with a bounding box, and then classifies the image patch within this bounding box~\cite{girshick2015fast,ren2015faster}.  
Nevertheless, this process can be streamlined with  a one-stage localisation and classification approach~\cite{lin2018focal} -- we explore this one-stage method in this paper~\cite{tian2019oneshot}.  

The classification of polyps typically focuses on solving problems containing two classes~\cite{hewett2012validation,komeda2017computer}, three classes~\cite{hayashi2013endoscopic,ribeiro2017exploring}, or four classes~\cite{iwatate2018validation}, while our approach is one of the first to study the five-class polyp classification problem~\cite{singh2013narrow,pu2018sa1908,tian2019oneshot}.
This five-class classification is claimed to be more effective than the other alternatives since it can enable colonoscopists to assess if a detected polyp is endoscopically resectable (i.e. pre-cancerous or early cancerous lesions -- classes $IIo$, $II$ and $IIIa$) or not endoscopically resectable (i.e., benign or invasive cancer - classes $I$ or $IIIb$, where for the latter class, the case is referred to surgery).  
Also, this classification can reduce costs and complications associated with polypectomy since the follow-up interval and method for endoscopic resection can differ depending on the number, size and type of the lesions found during the exam~\cite{levin2008screening,rex2017colorectal}. 
These advantages come with the drawback that this five-class classification tends to be more challenging for the classification model.

\subsubsection{Uncertainty and Calibration}

Deep learning models are generally trained with maximum likelihood estimation (MLE) that
tends to produce over-confident classifiers 
that represent poorly its expected accuracy~\cite{guo2017calibration}.  
Apart from a few papers~\cite{jiang2011calibrating,bullock2018xnet,eaton2018towards}, this issue has been overlooked by the medical image analysis community, even though it is an important feature in the deployment of systems in clinical settings.
Furthermore, MLE training cannot provide an estimation of uncertainty, unless through classification entropy~\cite{settles2012active} or some specific types of loss functions~\cite{ziyin2019simple}.
The use of Bayesian learning can provide a more reliable uncertainty estimation~\cite{gal2016dropout,gal2017concrete,kendall2017uncertainties} by considering that the model and observations are affected by noise processes. Such a learning process tends to have high training and testing computational costs~\cite{gamerman2006markov,jaakkola2000bayesian}, but recent work by Gal et al.~\cite{gal2016dropout,gal2017concrete,kendall2017uncertainties} addressed this issue with the proposal of efficient methods.
In fact, this method has been recently explored by medical image analysis methods~\cite{eaton2018towards,nair2018exploring}.
We argue that classification calibration and uncertainty estimation are two independent ideas and show that their combination can lead to effective classification strategies~\cite{carneiro2020deep}.

\subsubsection{Commercial Systems}

Recently, several companies have deployed commercially viable systems that integrate computer-aided detection and/or classification directly to the colonoscope output in real-time. 
For instance, polyp detection and classification systems have been developed by Olympus© in collaboration with with ai4gi©, by Fujifilm© with the CAD EYE$^{\text{TM}}$ technology, and by Medtronic© through GI GENIUS$^{\text{TM}}$. 
These systems tend to produce reliable polyp detection results, but the classification module does not seem to be integrated well with the detection module and appears to only separate the lesions into neoplastic (i.e. cancerous or precancerous) or non-neoplastic. Moreover, it is unclear if any of these systems can classify serrated lesions.
The classification of serrated lesions into neoplastic (i.e. traditional serrated adenomas and sessile serrated lesions/adenomas/polyps) and non-neoplastic (i.e. hyperplastic polyp) is clinically relevant and often difficult even for experienced endoscopists.

\section{Materials and Methods}
\label{sec:dataset_methods}


\subsection{Data Sets}
\label{sec:datasets}

We use four data sets for testing the proposed methods.  One is for the detection of colonoscopy frames containing polyps, and the other two are for the localisation and classification of polyps within an image that contains a polyp, where uncertainty and calibration are tested in one of these two data sets. 
These three data sets contain images and clinical information approved by the Human Research Ethics Committee (TQEH/LMH/MH/2008128 and HREC/16/TQEH/283) in Australia and by the Nagoya University Hospital Ethics Review Committee (2015-0485) in Japan. 
The fourth data set is the publicly available data set MICCAI 2015 Endoscopic Vision Challenge~\cite{bernal2017comparative} to test the localisation of polyps. 

\subsubsection{Polyp Detection}
\label{sec:polyp_detection_dataset}

\begin{table}[]
\centering
\caption{The training and testing sets used for polyp detection.}
\label{tab:polyp_detection_dataset}
\resizebox{\columnwidth}{!}{%
\begin{tabular}{@{}lrrrrr@{}}
\toprule
 & Patients & Normal & Abnormal & Water\&Feces & Total Images \\ 
 \hline 
Training set & 13 & 13250 & 100 & 3302 & 16652 \\ 
Testing set & 4 & 700 & 212 & 500 & 1612 \\ 
\bottomrule
\end{tabular}%
}
\end{table}

\begin{figure*}[t]
\begin{center}
\includegraphics[width=1.0\textwidth]{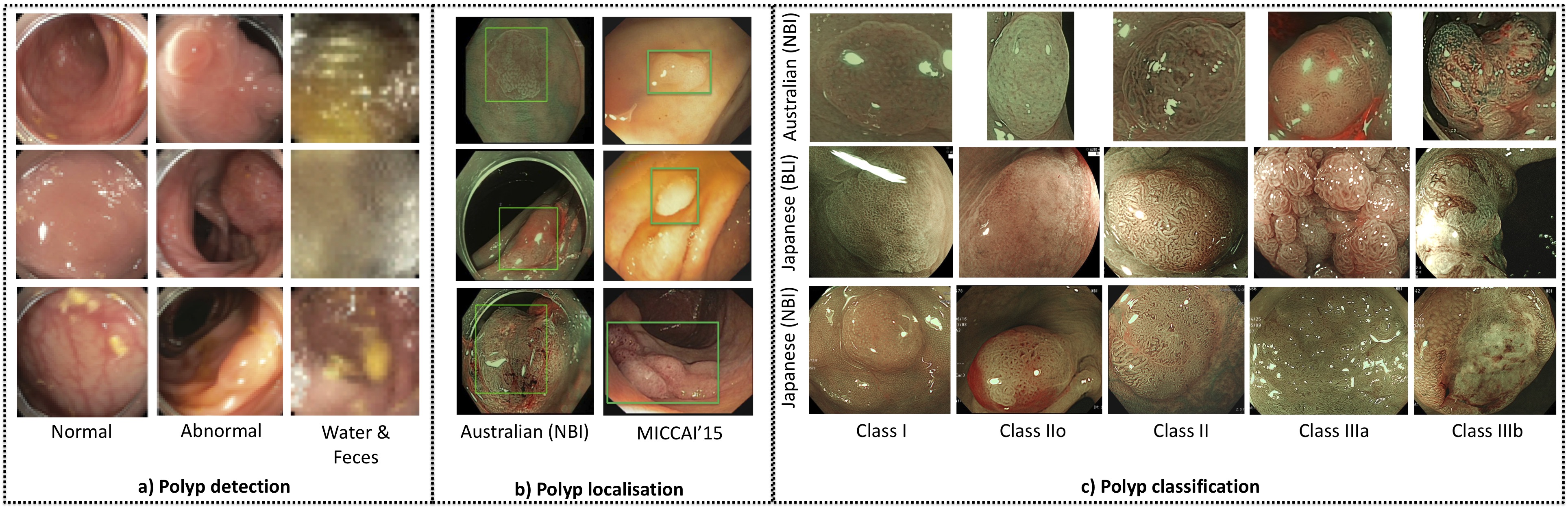} 
\end{center}
\caption{Data sets used in this chapter: a) polyp detection uses frames, labelled as Normal (showing healthy tissue), Abnormal (containing polyp), and Water\&Feces; b) polyp localisation from an image containing a polyp (the annotation is represented by a bounding box around the polyp); and c) polyp classification using the image patch inside the polyp bounding box.}
\label{fig:datasets}
\end{figure*}

For the detection of frames containing polyps, we used  colonoscopy videos captured with the Olympus~\textregistered 190 dual focus endoscope from 17 patients.  
We automatically remove frames with blurred visual information using the variance of Laplacian method~\cite{he2006laplacian}.
Next, we take one of every five consecutive frames to prevent the modelling of correlation between frames that are too close in time domain.
Table~\ref{tab:polyp_detection_dataset} shows details of the training and testing sets. 
This data set is represented by $\mathcal{D}^{(d)} = \{ \mathbf{x}_i ,t_i ,y_i \}_{i=1}^{|\mathcal{D}^{(d)}|}$, where $\mathbf{x}:\Omega \rightarrow \mathbb{R}^3$ denotes an RGB colonoscopy frame ($\Omega$ represents the frame lattice), $t_i \in \mathbb{N}$ represents patient identification\footnote{Note that the data set has been de-identified, so $t_i$ is useful only for splitting $\mathcal{D}^{(d)}$ into training, testing and validation sets in a patient-wise manner.}, and $y \in \mathcal{Y}^{(d)} = \{ \text{Normal, Polyp,  Water\&Feces} \}$ denotes the label.  In particular, the class $\text{Normal}$ represents frames in the set $\mathcal{D}^{(d)}_N \subset \mathcal{D}^{(d)}$, displaying healthy colon tissue; $\text{Abnormal}$ denotes frames in $\mathcal{D}^{(d)}_A \subset \mathcal{D}^{(d)}$, showing polyps; and 
$\text{Water\&Feces}$ in $\mathcal{D}^{(d)}_W \subset \mathcal{D}^{(d)}$, are frames containing water jet sprays and feces.
The patients in the testing, training and validations sets are mutually exclusive and the proportion of abnormal samples in the testing set is the typical proportion defined for other anomaly detection papers~\cite{perera2019ocgan,schlegl2019f}.  Samples of this data set are shown in Fig.~\ref{fig:datasets}-(a).

\subsubsection{Polyp Localisation and Classification (with Uncertainty and Calibration)}
\label{sec:polyp_localisation_classification_dataset}

For the localisation and classification of polyps, we rely on the data set defined by $\mathcal{D}^{(c)} = \{ \mathbf{x}_i,t_i,\mathbf{b}_i,y_i \}_{i=1}^{|\mathcal{D}^{(c)}|}$, where $\mathbf{x}$ and $t_i$ are as defined in Sec.~\ref{sec:polyp_detection_dataset}, $\mathbf{b}_i \in \mathbb R^4$ denotes the two 2-D coordinates of the bounding box containing the polyp,
and $y_i \in \mathcal{Y}^{(c)} = \{ I, II, IIo, IIIa, IIIb \}$ denotes the five-polyp classification~\cite{singh2013narrow,pu2018sa1908,tian2019oneshot}, divided into:  1) hyperplastic polyp ($I$), 2) sessile serrated lesion ($IIo$), 3) low grade tubular adenoma ($II$), 4) high grade adenoma/ tubulovillous adenoma /superficial cancer ($IIIa$), and 5) invasive cancer ($IIIb$) -- see Fig.~\ref{fig:datasets}-(b),(c).
Figure~\ref{fig:datasets}-(b) also shows samples of the MICCAI 2015 Endoscopic Vision Challenge data set~\cite{bernal2017comparative}.

The data set $\mathcal{D}^{(c)}$ contains images collected from two different sites, referred to as Australian and Japanese sites, which allow us to test our classification method in a more realistic experimental set-up, where the training and testing are from different domains -- see Fig.~\ref{fig:datasets}-(c).
The Australian data set contains images of colorectal polyps collected from a tertiary hospital in South Australia with the Olympus~\textregistered 190 dual focus colonoscope using Narrow Band Imaging (NBI), and is represented by $\mathcal{D}_A^{(c)} \subset \mathcal{D}^{(c)}$. 
The number of images of $\mathcal{D}_A^{(c)}$ is $871$, which were scanned from $218$ patients, where $102$ images from $39$ patients are of class $I$, $346$ images from $93$ patients are of class $II$, $281$ images from $48$ patients are of class $IIo$, $79$ images from $25$ patients are of class $IIIa$ and $63$ images from 14 patients are of class $IIIb$.
The Japanese data set, denoted by $\mathcal{D}_J^{(c)} \subset \mathcal{D}^{(c)}$, contains two subsets of colorectal polyp images acquired from a tertiary hospital image database in Japan: magnified NBI images obtained from the Olympus~\textregistered 290 series, and magnified Blue Laser Imaging (BLI) images from the Fujifilm~\textregistered 700 series. 
The Japanese data set has $20$ NBI images from $20$ patients and $49$ BLI images from $49$ patients, where the NBI set has $3$ images of class $I$, $5$ images of class $II$, $2$ images of class $IIo$, $7$ images of class $IIIa$ and $3$ images of class $IIIb$.  
The BLI data set contains $9$ images of class $I$, $10$ images of class $II$, $10$ images of class $IIo$, $11$ images of class $IIIa$ and $9$ images of class $IIIb$.
All images from the Australian and Japanese data sets were correlated with histology and de-identified into folders according to the five classes defined above~\cite{singh2013narrow,pu2018sa1908,tian2019oneshot}. 

\begin{figure}[t]
\begin{center}
\includegraphics[width=0.49\textwidth]{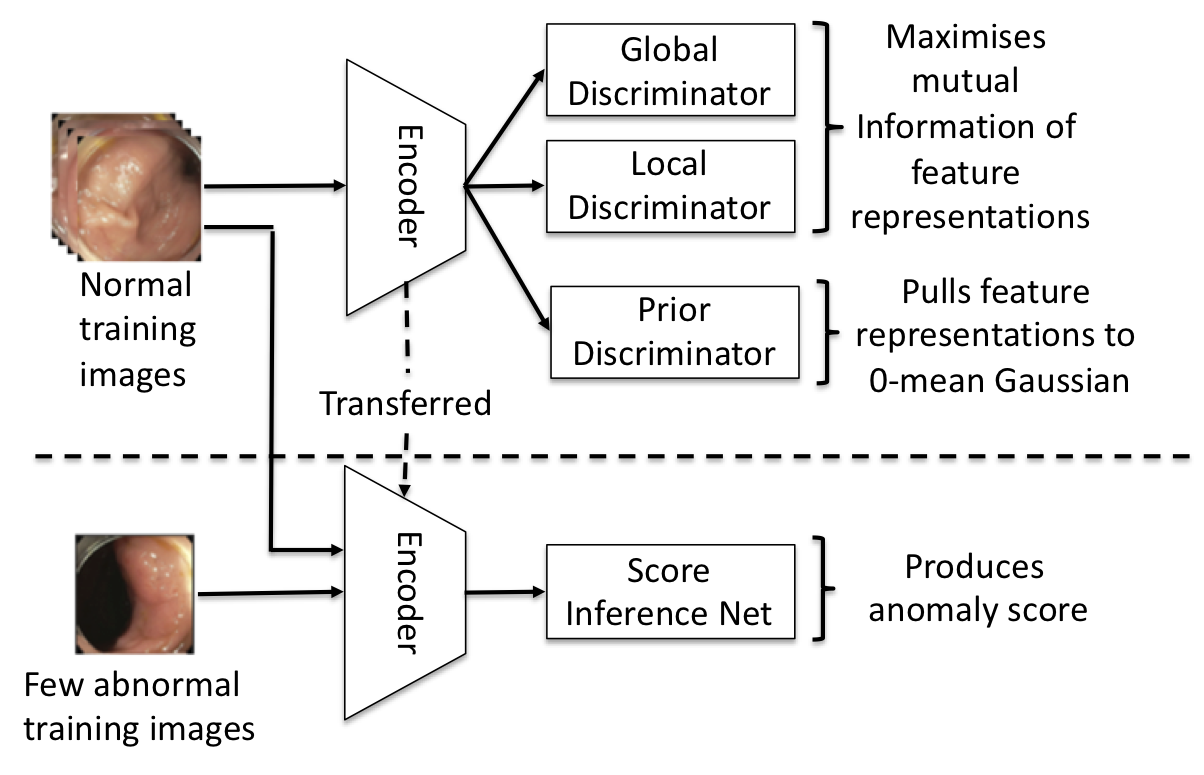} 
\end{center}
\caption{Diagram of the few-shot anomaly detector FSAD-NET with the representation learning (top) and SIN learning (bottom).}
\label{fig:anomaly_detection}
\end{figure}

The MICCAI 2015 Endoscopic Vision Challenge data set~\cite{bernal2017comparative}, represented by $\mathcal{D}_M^{(l)}$, used for the polyp localisation experiment, comprises the CVC-CLINIC and ETIS-LARIB data sets. 
CVC-CLINIC has 612 images (size  $388\times284$ pixels) from 31 different
polyps and 31 sequences (captured with Olympus
Q160AL and Q165L, Exera II video-processor). ETIS-LARIB has
196 frames (size $1225\times966$) from 44 different
polyps and 34 sequences (captured with Pentax 90i series, EPKi 7000
video-processor). 
All images in this MICCAI 2015 data set have at least one polyp annotated by
expert endoscopists.

We consider three experiments: 1) training and testing from the MICCAI 2015 data set, 2) training and testing images are from the Australian data set, and 3) training from the Australian data set and testing from Japanese data set.
For the MICCAI 2015 data set, we folow the experimental setup described in~\cite{bernal2017comparative}.
For the experiment based on training and testing sets from the Australian data set, the training set  $\mathcal{T}^{(c)}$ has images from $60\%$ of the patients, the validation set $\mathcal{V}^{(c)}$ has  images from $20\%$ of the patients, and the testing set has images from the remaining $20\%$ of the patients, where the patients in the testing, training and validations sets are mutually exclusive and each one of these subsets are randomly formed to have a similar proportion of the five classes.
For the experiment based on training on the Australian data set and testing on the Japanese data set, the testing set contains only images from the Japanese site.
The Australian data set is used for training and testing the proposed method using a cross validation experiment, while the Japanese data set is used exclusively for testing the system, enabling us to test the performance of the method with images collected from different colonoscopes (Olympus~\textregistered 190 and 290 series) and different technologies (i.e., NBI and BLI).

\subsection{Methods}
\label{sec:methods}

In this section, we present the binary classifier that rejects frames with water jet sprays and feces, then, we describe
the few-shot anomaly detection method for  identifying frames with polyps. Next, we explain the method for localising and classifying polyps, followed by the approach to calibrate confidence and estimate classification uncertainty.

\subsubsection{Detection of Frames with Water Jet Sprays and Feces}
\label{sec:method_detection_water_feces}

This first stage consists of a model, trained with binary cross entropy (BCE) loss~\cite{bishop2006pattern}, which classifies images belonging to the Water\&Feces class versus images belonging to the Normal and Abnormal classes, using the data set in Table~\ref{tab:polyp_detection_dataset}.  This model is represented by
\begin{equation}
    D(y | \mathbf{x},\theta_{D}),
    \label{eq:model_reject_water_feces}
\end{equation}
where $y \in \{\text{Normal OR Abnormal, Water\&Feces} \}$, and $\theta_{D}$ represents the model parameters.

\subsubsection{Few-shot Polyp Detection}
\label{sec:method_few_shot_polyp_detection}

Anomaly detection approaches are usually designed as a one-class classifier or a few-shot learning method~\cite{pang2020deep}.
Results from approaches exploring these two ideas~\cite{liu2020photoshopping,tian2020few} suggest that the latter is more effective, so we focus on it and propose FSAD-NET~\cite{tian2020few}, which consists of a feature encoder and a score inference network (SIN), as displayed in Fig.~\ref{fig:anomaly_detection}.  The training process is divided into two stages: 1)  training of the feature encoder $\mathbf{e} = E(\mathbf{x};\theta_E)$ ($\theta_E$ represents the encoder parameter and $\mathbf{e} \in \mathbb R^E$ denotes the embedding) to maximise the mutual information (MI) between normal images $\mathbf{x} \in \mathcal{D}^{(d)}_N$ and their embeddings $\mathbf{e}$~\cite{hjelm2018learning}; and 2) training of the SIN $S(E(\mathbf{x};\theta_E);\theta_S)$~\cite{pang2019deep} ($\theta_S$ represents the SIN parameter), with a contrastive-like loss that relies on $\mathcal{D}^{(d)}_N$ and $\mathcal{D}^{(d)}_A$ to reach the following condition: 
$ S(E(\mathbf{x}\in \mathcal{D}^{(d)}_A; \theta_E);\theta_S) >  S(E(\mathbf{x}\in \mathcal{D}^{(d)}_N;\theta_E);\theta_S) $.    

The MI maximisation in step one is achieved with~\cite{hjelm2018learning}:
\begin{equation}
\begin{split}
\theta_E^*,\theta_D^*,\theta_L^* = \arg & \max_{\theta_E,\theta_D,\theta_L} \Big ( 
\alpha \hat{I}_{\theta_D}( \mathbf{x} ; E(\mathbf{x};\theta_E)) \\
&+ \frac{\beta}{|\mathcal{M}|} \sum_{\omega \in \mathcal{M}} \hat{I}_{\theta_L}( \mathbf{x}(\omega) ; E(\mathbf{x}(\omega);\theta_E)  ) \Big ) \\
& + \gamma \arg\min_{\theta_E}\arg\max_{\phi}  \hat{F}_{\phi} (\mathbb{V} || \mathbb{U}_{\mathbb{P},\theta_E}),
\end{split}
   \label{eq:train_E}
\end{equation}
where $\alpha,\beta,\gamma$ denote model hyperparameters, $\hat{I}_{\theta_D}(.)$ and $\hat{I}_{\theta_L}(.)$ represent an MI lower bound based on the Donsker-Varadhan representation of the Kullback-Leibler (KL)-divergence~\cite{hjelm2018learning}, defined by 
\begin{equation}
\begin{split}
    \hat{I}_{\theta_D}(\mathbf{x}; &  E(\mathbf{x};\theta_E)) =  \\
    & \mathbb{E}_{\mathbb{J}}[D(\mathbf{x},E(\mathbf{x};\theta_E);\theta_D)]-\log \mathbb{E}_{\mathbb{M}}[e^{D(\mathbf{x},E(\mathbf{x};\theta_E);\theta_D)}],
    \end{split}
    \label{eq:MI}
\end{equation}
with $\mathbb{J}$ denoting the joint distribution between images and their embeddings, $\mathbb{M}$ representing the product of the marginals of the images and embeddings, and $D(\mathbf{x},E(\mathbf{x};\theta_E);\theta_D)$ being a discriminator with parameter $\theta_D$. 
The function $\hat{I}_{\theta_L}( \mathbf{x}(\omega) ; E(\mathbf{x}(\omega);\theta_E))$ in~\eqref{eq:train_E} has a similar definition as~\eqref{eq:MI} for the discriminator $L(\mathbf{x}(\omega),E(\mathbf{x}(\omega);\theta_E);\theta_L)$, and represents the local MI between image regions $\mathbf{x}(\omega)$ ($\omega \in \mathcal{M} \subset \Omega$, with $\mathcal{M}$ denoting a sub-set of the image lattice) and respective local  embeddings $E(\mathbf{x}(\omega),\theta_E)$. 
Also in~\eqref{eq:train_E},
\begin{equation}
\begin{split}
\arg\min_{\theta_E} & \arg\max_{\theta_F} \hat{F}_{\phi} (\mathbb{V} || \mathbb{U}_{\mathbb{P},\theta_E}) = \\ & \mathbb{E}_{\mathbb V}[\log F(\mathbf{e};\theta_F)] + \mathbb{E}_{\mathbb{P}}[\log(1-F(E(\mathbf{x};\theta_E));\theta_F))],
\end{split}
\label{eq:div_E}
\end{equation}
where $\mathbb{V}$ represents a prior distribution for the embeddings $\mathbf{e}$ (we assume $\mathbb{V}$ to be a normal distribution $\mathcal{N}(.;\mu_{\mathbb{V}},\Sigma_{\mathbb{V}})$, with mean $\mu_{\mathbb{V}}$ and covariance $\Sigma_{\mathbb{V}}$), $\mathbb{P}$ is the distribution 
of the embeddings $\mathbf{e}=E(\mathbf{x}\in\mathcal{D}^{(d)}_N;\theta_E)$, and $F(.;\theta_F)$ represents a discriminator learned with adversarial training to estimate the probability that the input is sampled from either $\mathbb{V}$ or $\mathbb{P}$.  
This objective function attracts the normal image embeddings toward $\mathcal{N}(.;\mu_{\mathbb{V}},\Sigma_{\mathbb{V}})$.

The SIN training takes the embeddings of normal and abnormal images with $\mathbf{e} = E(\mathbf{x} \in \mathcal{D}^{(d)}_A \bigcup \mathcal{D}^{(d)}_N;\theta_E^*)$ to learn $ \theta_S^*$ for $S(.)$ using the contrastive-like loss~\cite{pang2019deep}
\begin{equation}
\begin{split}
    \ell_S = & \mathbb{I}(y\textrm{ is }Normal)|T(S(\mathbf{z};\theta_S);\mu_T,\sigma_T)| + \\
    & \mathbb{I}(y\textrm{ is }Abnormal)\max(0,a - T(S(\mathbf{e};\theta_S);\mu_T,\sigma_T)),
    \end{split}
    \label{eq:loss_score_inference_net}
\end{equation}
where $\mathbb{I}(.)$ denotes the indicator function, $T(x;\mu_T,\sigma_T) = \frac{x - \mu_T}{\sigma_T}$ with $\mu_T=0$ and $\sigma_T=1$ representing the mean and standard deviation of the prior distribution for the anomaly scores for normal images, and $a$ is the margin between $\mu_T$ and the anomaly scores of abnormal images~\cite{pang2019deep}.  The loss in~\eqref{eq:loss_score_inference_net} pulls the scores from normal images to $\mu_T$ and pushes the scores of abnormal images away from $\mu_T$ with a minimum margin of $a$.

The inference consists of taking a test image $\mathbf{x}$, estimating the embedding with $\mathbf{e} = E(\mathbf{x};\theta_E^*)$ and computing the score with $s = S(\mathbf{e};\theta_S^*)$.  We then compare this score $s$ to a threshold $\tau$ to estimate if the test image is normal or abnormal.

\subsubsection{Localisation and Classification of Polyps}
\label{sec:localisation_classification_polyps}

\begin{figure}[t]
\begin{center}
\includegraphics[width=0.49\textwidth]{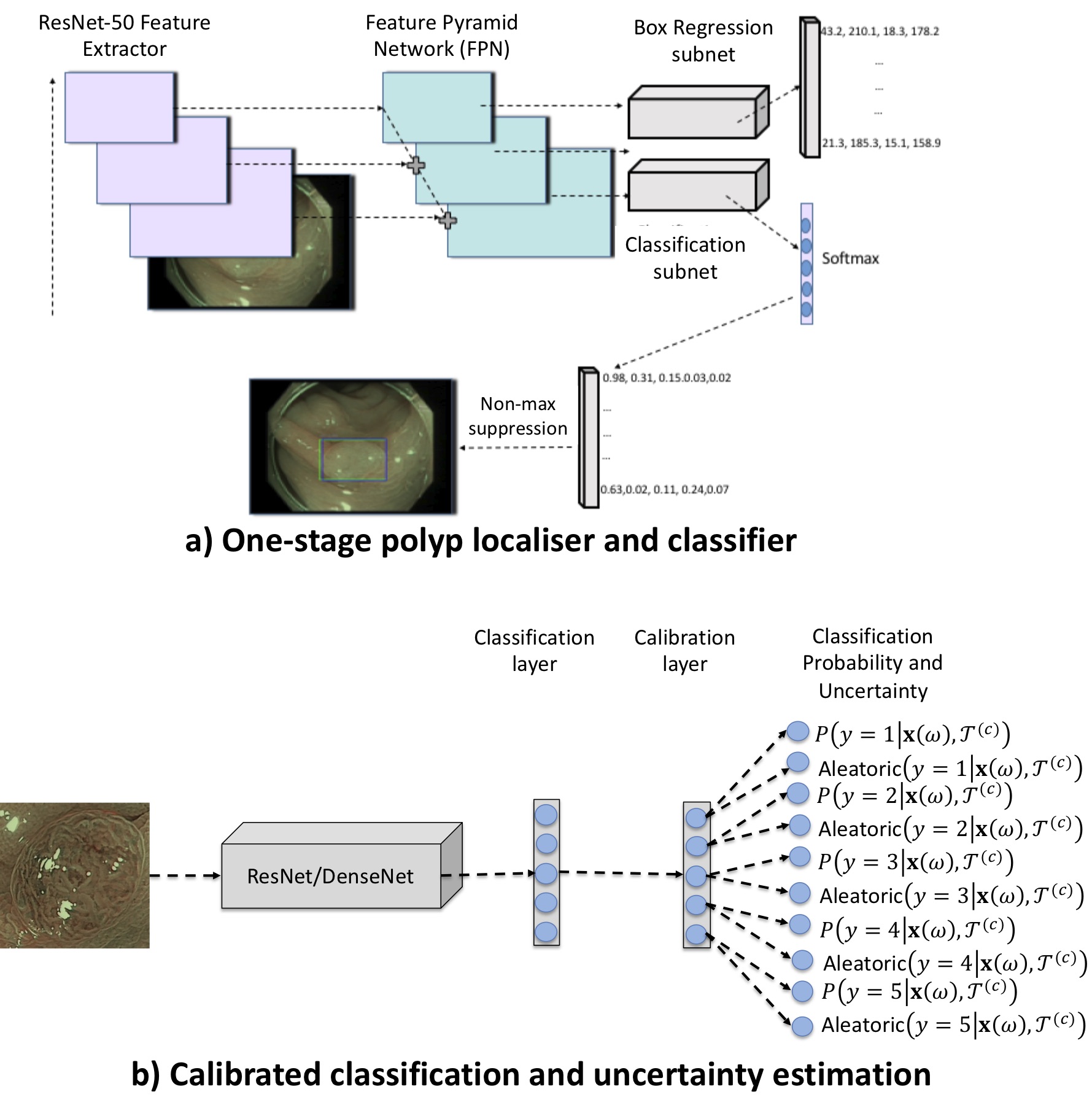} 
\end{center}
\caption{Diagrams of the one-stage polyp localiser and classifier (a) and the calibrated classification and uncertainty estimation model (b).}
\label{fig:classification_localisation_models}
\end{figure}

Our approach explores the efficient
method that can detect and classify polyps simultaneously~\cite{lin2018focal,tian2019oneshot}.
This one-stage model, depicted in Fig.~\ref{fig:classification_localisation_models}-(a), consists of a feature pyramid network (FPN)~\cite{lin2017feature}, followed by a regressor that outputs a fixed list of bounding boxes and a classifier that returns a 5-class classification probability vector for each bounding box. 
This model is represented by~\cite{lin2018focal}:
\begin{equation}
\mathbf{b},\mathbf{p} = C(\mathbf{x},\theta_C),  
\label{eq:classifier_localiser}
\end{equation}
where $\mathbf{b}\in\mathbb R^4$ represents the detected bounding box and $\mathbf{p} \in [0,1]^{|\mathcal{Y}^{(c)}|}$ denotes the 5-class classification probability vector of the image patch.

The training for the polyp localisation in~\eqref{eq:classifier_localiser} assumes only two classes for a bounding box: normal (background) or polyp of any class (foreground), and we rely on the focal loss to train the polyp localisation~\cite{lin2018focal}: 
\begin{equation}
\ell_F(p) = - \alpha(1 - p)^{\gamma}log(p),
\label{eq:focal_loss}
\end{equation}
where $p = P(y=\text{foreground}|\mathbf{x}(\mathcal{M}),\theta_C)$ if the image patch $\mathbf{x}(\mathcal{M})$ ($\mathcal{M} \subset \Omega$) contains a polyp, $p = 1-P(y=\text{foreground}|\mathbf{x}(\mathcal{M}),\theta_C)$, otherwise, with $P(y=\text{foreground}|\mathbf{x}(\mathcal{M}),\theta_C) \in [0,1]$ denoting the model's probability that the image patch $\mathbf{x}(\mathcal{M})$ shows a polyp,  $\gamma \in [0,5]$ modulates the loss for well-classified image patches, and $\alpha \in [0,1]$ denotes a hyper-parameter for the class weight (foreground or background) of the image patch $\mathbf{x}(\mathcal{M})$.  The decision of whether an image patch contains a polyp is based on an intersection over union (IoU) larger than a threshold $\tau$ between $\mathbf{x}(\mathcal{M})$ and the patch formed by the manual annotation for the bounding box $\mathbf{b} \in \mathbb R^4$ present in $\mathcal{D}^{(c)}$, defined in Sec.~\ref{sec:polyp_localisation_classification_dataset}.  
The training for the polyp classification part of the model in~\eqref{eq:classifier_localiser} is based on the bounding boxes that have an IoU larger than a threshold $\tau$ and a confidence score for each class. 
To reduce the number of the bounding boxes, 
we merge the ones with same classes and IoU larger than 0.5 and assigned with the maximum confidence score . 
To represent each image with only one class, we first select the bounding box with the maximum confidence score as the location of the polyp but omit the class label. For each class, we compute the sum of the scores on all bounding boxes as the final confidence score of that class. Then, the class is  determined by the maximum confidence score. 
The classification model is trained with multi-class cross entropy loss.

During inference, the bounding boxes with confidence scores larger than a threshold $\tau = 0.05$ and IoU $>0.5$ are merged with non-maximum suppression, where the confidence of the merged bounding boxes is represented by the sum of the confidences 
of the merged bounding boxes.  The final classification is obtained from the maximum confidence.

\subsubsection{Polyp Classification Uncertainty and Calibration}
\label{sec:uncertainty_calibration}

For the model in Fig.~\ref{fig:classification_localisation_models}-(b) that produces a calibrated confidence and uncertainty estimation, we assume that the input is an image patch $\mathbf{x}(\mathcal{M})$ (sampled to have a fixed size) inside the bounding box $\mathbf{b}$ that contains the polyp.  This model estimates the probability of each class $y \in \mathcal{Y}^{(c)}$ with
\begin{equation}
P(y | \mathbf{x}(\mathcal{M}),\theta_P),
\label{eq:basic_model}
\end{equation} 
where $\theta_P$ parameterises the model.
To allow the estimation of classification uncertainty, we rely on a Bayesian approach,
where the goal is to
estimate a distribution for $\theta_P$ (with prior $P(\theta_P)$ and posterior $P(\theta_P | \mathcal{T}^{(c)})$), and inference is processed with~\cite{carneiro2020deep,gal2016dropout}:
\begin{equation}
P(y | \mathbf{x}(\mathcal{M}), \mathcal{T}^{(c)}) = \int_{\theta_P} P(y | \mathbf{x}(\mathcal{M}),\theta_P)P( \theta_P | \mathcal{T}^{(c)}) d\theta_P.
\label{eq:BE}
\end{equation} 
The inference in~\eqref{eq:BE} is generally intractable, an issue that can be mitigated with the use of a variational distribution $Q(\theta_P | \psi_Q)$ to approximate $P(\theta_P | \mathcal{T}^{(c)})$, where $\psi_Q = \{ \mathbf{M}_l, p_l \}_{l=1}^L$, with $L$ denoting the number of layers in the model, $\mathbf{M}_l$ representing the layer-wise mean weight matrices and $p_l$ the dropout probabilities. 
This means that $Q(\theta_P| \psi_Q) = \prod_l \mathbf{M}_l \times \text{diag}(\text{Bernoulli}(1-p_l)^{K_l})$, with the weight matrix in layer $l$ having dimensions $K_{l+1} \times K_l$~\cite{gal2016dropout,gal2017concrete}.  Then, Eq.~\ref{eq:BE} can be solved with Monte Carlo (MC) integration: 
\begin{equation}
\begin{split}
Q(y | \mathbf{x}(\mathcal{M}), \psi_Q) & = \frac{1}{N} \sum_{j=1}^N P(y | \mathbf{x}(\mathcal{M}), \hat{\theta}_{Pj}) \\
& = \frac{1}{N} \sum_{j=1}^N \sigma(f^{\hat{\theta}_{Pj}}(\mathbf{x}(\mathcal{M}))),
\end{split}
\label{eq:BE_inference}
\end{equation}
where $ \hat{\theta}_{Pj} \sim Q(\theta_P| \psi_Q)$, $\sigma(.)$ represents the softmax function, and $f^{\hat{\theta}_{Pj}}(\mathbf{x}(\mathcal{M})) \in \mathbb R^{|\mathcal{Y}^{(c)}|}$ is the logit vector for the final softmax function applied by the classifier.  The learning of the variational distribution parameter $\psi_Q^*$ is based on the minimisation of the following loss~\cite{gal2016dropout,gal2017concrete}:
\begin{equation}
\begin{split}
\ell(\psi_Q) = &  -\int Q(\theta_P| \psi_Q)  \log \prod_{i=1}^{|\mathcal{T}^{(c)}|} P( y_i | \mathbf{x}_i(\omega), \theta_P) d\theta_P + \\
 & KL(Q(\theta_P| \psi_Q) || P(\theta_P)),
 \end{split}
\label{eq:BE_train}
\end{equation}
where $KL(.)$ represents the Kullback-Leibler divergence.
The integral in~\eqref{eq:BE_train} is approximated with MC integration,
\begin{equation}
\begin{split}
\ell(\psi_Q) \approx & - \frac{1}{N} \sum_{j=1}^N \log \prod_{i=1}^{|\mathcal{T}^{(c)}|} P( y_i | \mathbf{x}_i(\omega), \hat{\theta}_{Pj}) + \\ & KL(Q(\theta_P| \psi_Q) || P(\theta_P)),  
\end{split}
\label{eq:BE_train_MC}
\end{equation}
with $\hat{\theta}_{Pj} \sim Q(\theta_P| \psi_Q)$. 
As explained in~\cite{gal2017concrete}, we assume that $P(\theta_P)$ has a prior distribution represented by a discrete quantised Gaussian prior that enables an analytically derivation of $KL(Q(\theta_P| \psi_Q) || P(\theta_P))$ in~\eqref{eq:BE_train}.
As depicted in Fig.~\ref{fig:classification_localisation_models}-(b), the training process, based on concrete dropout~\cite{gal2017concrete}, produces a model that outputs the classification probability and the aleatoric uncertainty for each class~\cite{kendall2017uncertainties}.  
The estimation of classification uncertainty can then be performed in many ways, such as based on the entropy of the probability vector~\cite{settles2012active,kendall2017uncertainties}, computed with~\cite{carneiro2020deep}:
\begin{equation}
\begin{split}
H&(P(y|\mathbf{x}(\mathcal{M}),\mathcal{T}^{(c)})) = \\ & -\sum_{c \in \mathcal{Y}^{(c)}} P(y=c|\mathbf{x}(\mathcal{M}),\mathcal{T}^{(c)}) \log (P(y=c|\mathbf{x}(\mathcal{M}),\mathcal{T}^{(c)})),
\end{split}
\label{eq:entropy}
\end{equation}
where $P(y|\mathbf{x}(\mathcal{M}),\mathcal{T}^{(c)})$ can be replaced by $P( y | \mathbf{x}(\mathcal{M}), \theta_P^*)$ from~\eqref{eq:basic_model}, $Q(y|\mathbf{x}(\mathcal{M}), \psi^*_Q)$ from~\eqref{eq:BE_inference} or the calibrated classifiers from~\eqref{eq:temp_scaling}.

Another issue with the training above is that the classification result is likely to be over-confident~\cite{guo2017calibration}.  This is certainly undesirable in a clinical setting, but it can be fixed with confidence calibration~\cite{guo2017calibration} using a post-processing method that modifies the output classification probability computation as follows~\cite{carneiro2020deep}: 
\begin{equation}
\widetilde{Q}(y | \mathbf{x}(\mathcal{M}),s,\psi^*_Q) = \frac{1}{N} \sum_{j=1}^N \sigma(   \mathbf{f}^{\hat{\theta}_{Pj}}(\mathbf{x}(\mathcal{M}))/s   ),
\label{eq:temp_scaling}
\end{equation}
where $\hat{\theta}_{Pj} \sim Q(\theta_P| \psi^*_Q)$, $\mathbf{f}^{\hat{\theta_P}}(\mathbf{x}(\mathcal{M}))$ denotes the logit from~\eqref{eq:BE_inference}, and $s \in \mathbb R^+$ is a temperature parameter that smooths the softmax function $\sigma(.)$ by increasing its entropy and is learned with stochastic gradient descent using the validation set $\mathcal{V}^{(c)}$~\cite{guo2017calibration}.

Using uncertainty estimation and confidence calibration, we formulate a method that uses that information to improve classification accuracy.  In particular, we define two hyper-parameters $\tau^*_1(Z)$ and $\tau^*_2(Z)$, learned from the validation set $\mathcal{V}^{(c)}$ to enable the rejection of a percentage $Z$ of samples that have low confidence and high uncertainty, respectively as follows:
\begin{equation}
\begin{split}
1) & \;\;\; P(y | \mathbf{x}(\mathcal{M}), \mathcal{T}^{(c)}) < \tau^*_1(Z)\text{, and}\\
2) & \;\;\; H(Q(y|\mathbf{x}(\mathcal{M}),\psi^*_Q)) > \tau^*_2(Z),
\end{split}
\label{eq:rejecting_testing}
\end{equation}
where the thresholds are learned with 
$\tau^*_1(Z) =  P_{sorted}(Z \times |\mathcal{V}^{(c)}|)$ and $\tau^*_2(Z) =  H_{sorted}(Z \times |\mathcal{V}^{(c)}|)$.
$P_{sorted}$ has the values of $\max_{y \in\mathcal{Y}^{(c)}} P(y | \mathbf{x}(\mathcal{M}), \mathcal{T}^{(c)}) $ sorted in ascending order for  the samples in the validation set $\mathcal{V}^{(c)}$, and $H_{sorted}$ has the values of $H(Q(y|\mathbf{x}(\mathcal{M}),\psi^*_Q))$, defined in~\eqref{eq:entropy}, sorted in descending order for the samples in $\mathcal{V}^{(c)}$. 

\section{Results and Discussion}
\label{sec:experiments}

In this section, we first present the experiments that test the polyp detection methods, followed by another set of experiments on polyp localisation and classification, and final experiments on confidence calibration and uncertainty estimation.

\subsection{Polyp Detection Experiments}

We test the polyp detection method under two assumptions. We first present the results of FSAD-NET from Sec.~\ref{sec:method_few_shot_polyp_detection} working under the assumption that there are no frames containing water and feces -- this means that we use the training and testing set from Tab.~\ref{tab:polyp_detection_dataset} without the Water\&Feces frames.  We then show FSAD-NET results using all sets in Tab.~\ref{tab:polyp_detection_dataset}, where the Water\&Feces frames are rejected by the binary classifier in~\eqref{eq:model_reject_water_feces}.

For these two experiments, 
the original colonoscopy image had the original resolution of 1072 $\times$ 1072 $\times$ 3 reduced to 64 $\times$ 64 $\times$ 3 to reach a good trade-off between computational cost (for training and testing) and detection accuracy.
For the proposed FSAD-NET, we use Adam~\cite{kingma2015adam} optimiser during training with a learning rate of 0.0001, where model selection (to estimate architecture, learning rate, mini-batch size and number of epochs) is done with the validation set $\mathcal{V}^{(d)}$.
The binary classifier to detect Water\&Feces frames is represented by a DenseNet~\cite{huang2017densely} trained from scratch with Adam optimiser~\cite{kingma2015adam} with a learning rate of 0.001 and mini-batch of 128 for 200 epoches.


For the few-shot anomaly detector FSAD-NET~\cite{tian2020few}, 
we use a similar backbone architecture as other competing approaches in Tab.~\ref{tab:result_polyp_detection}, where the encoder $E(.;\theta_E)$ uses four convolution layers (with 64, 128, 256, 512 filters of size 4 $\times$ 4), the global discriminator (used in $\hat{I}_{\theta_D}$ from Eq.~\ref{eq:train_E}) has three convolutional layers (with 128, 64, 32 filters of size 3 $\times$ 3), and the local discriminator (used in $\hat{I}_{\theta_L}$ from Eq.~\ref{eq:train_E})  has three convolutional layers (with 192, 512, 512 filters of size 1 $\times$ 1). 
The prior discriminator $F(.;\theta_F)$ in~\eqref{eq:div_E} has three linear layers with 1000, 200, and 1 node per layer.
We also use the validation set to estimate $a=6$ in~\eqref{eq:loss_score_inference_net}.
In~\eqref{eq:train_E}, we follow the DIM paper for setting the hyper-parameters with~\cite{hjelm2018learning}: $\alpha=0.5$, $\gamma=1$, $\beta=0.1$. For the prior distribution for the embeddings in~\eqref{eq:div_E}, we set $\mu_{\mathbb V}=\mathbf{0}$ (i.e., a $Z$-dimensional vector of zeros), and $\Sigma_{\mathbb V}$ is a $Z \times Z$ identity matrix.
To train the model, we first train the encoder, local, global and prior discriminator (they denote the representation learning stage) for 6000 epochs with a mini-batch of 64 samples. 
We then train SIN for 1000 epochs, with a batch size of 64, while fixing the parameters of encoder, local, global and prior discriminator.
We implement all methods above using Pytorch~\cite{paszke2017automatic}.

The detection results are measured with the area under the receiver operating characteristic curve (AUC) on the test set~\cite{schlegl2019f,perera2019ocgan}, computed by varying a threshold $\tau$ for the SIN score $S(\mathbf{e};\theta_S)$ in~\eqref{eq:loss_score_inference_net} for FSAD-NET.

Table~\ref{tab:result_polyp_detection} shows the results of several methods working under the assumption that there are no frames containing water and feces. 
In this table, the results of the proposed FSAD-NET~\cite{tian2020few} are compared with\footnote{Codes for the other methods were downloaded from the authors' Github pages and tuned for our problem.} other zero-shot and few-shot anomaly detectors.  The zero-shot detectors considered are: DAE~\cite{masci2011stacked} and VAE~\cite{doersch2016tutorial}, OCGAN~\cite{perera2019ocgan}, f-anoGAN and its variants~\cite{schlegl2019f} represented by image-to-image mean square error (MSE) loss (izi), Z-to-Z MSE loss (ziz) and its hybrid version (izif), and ADGAN~\cite{liu2020photoshopping}. 
For the few-shot approaches, we considered several methods trained with 40 polyp images and all normal images from Tab.~\ref{tab:polyp_detection_dataset} (40 polyp images is the number of training images needed to get stable results before reaching a diminishing return behaviour~\cite{tian2020few}), such as Densenet121~\cite{huang2017densely} trained with a large amount of data augmentation to deal with the extreme training imbalance, and several variants of the FSAD-NET~\cite{tian2020few}.
The variants 'Cross entropy' and 'Focal loss' replace the contrastive loss in~\eqref{eq:loss_score_inference_net} by the cross entropy loss (commonly used in classification problems)~\cite{goodfellow2016deep} and the focal loss (robust to imbalanced learning problems)~\cite{lin2018focal}, respectively. 
The method 'without RL' tests the representation learning (RL) role by removing it from the FSAD-NET formulation.  
The few-shot training algorithm 'Learning to Reweight'~\cite{ren2018learning} represents an example of a method designed to handle imbalanced learning and we use it for training FSAD-NET.
The importance of DIM to train the encoder is tested by replacing it with a deep auto-encoder~\cite{masci2011stacked} -- this is labelled as 'AE network'.

\begin{figure}[t!]
\small
\begin{center}
\includegraphics[width = 1.0\linewidth]{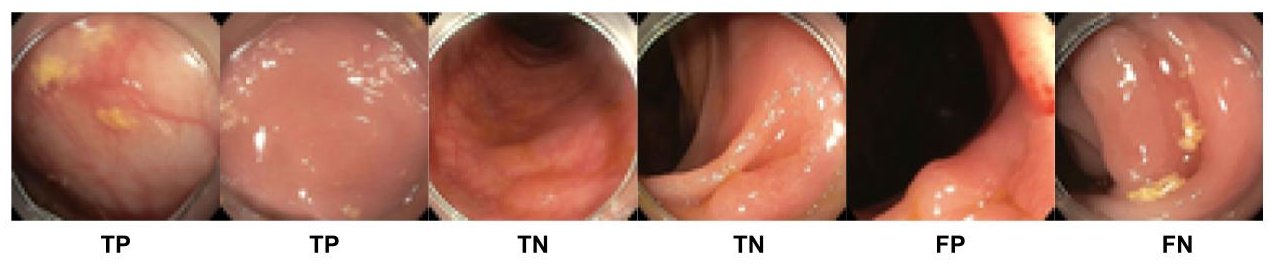}
\end{center}
\caption{Example of the FSDA-NET detection results, where TN denotes a successful detection of a normal frame, TP is a successful detection of an abnormal frame, FN denotes an unsuccessful detection of a normal frame and FP, an unsuccessful detection of an abnormal frame.}
\label{fig:results_anomaly_detection}
\end{figure}

The results in Table~\ref{tab:result_polyp_detection} show that 
few-shot anomaly detection is generally more accurate than zero-shot.
When comparing FSAD-NET with Densenet121~\cite{huang2017densely}, we notice that FSAD-NET is more accurate by a large margin.  
Results also demonstrate that the contrastive loss used in FSAD-NET is more appropriate for few-shot anomaly detection, compared with cross-entropy.
The removal of the representation learning reduces substantially the AUC result.
The use of Learning to Reweight~\cite{ren2018learning} for FSAD-NET shows relatively good results, with an AUC of 78.62\%, but they are not competitive with the final FSAD-NET results.
The results from the AE network show that FSAD-NET is more accurate, suggesting the effectiveness of using MI and prior distribution for learning the feature embeddings.
Figure~\ref{fig:results_anomaly_detection} shows examples of the FSAD-NET detection results.

\begin{figure}[t!]
\small
\begin{center}
\includegraphics[width = 0.9\linewidth]{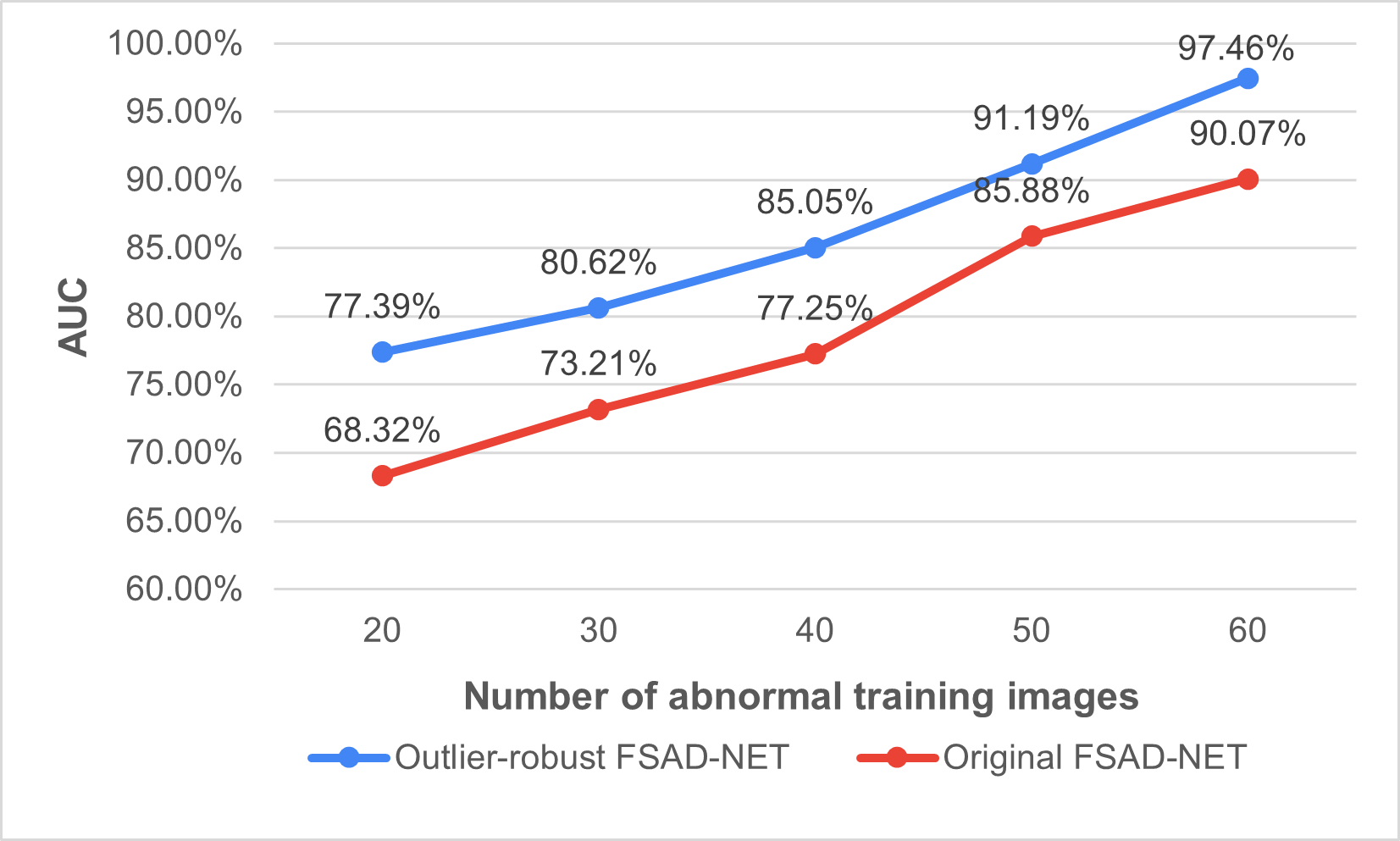}
\end{center}
\caption{Results on the data set that includes Water\&Feces frames.  Mean AUC results from our outlier-robust FSAD-NET as a function of the number of abnormal training images present in the training set, comparing with the original FSAD-NET model.}
\label{fig:results_outlier}
\vspace{-10pt}
\end{figure}

By rejecting Water\&Feces frames by the binary classifier in~\eqref{eq:model_reject_water_feces}, we build the 'outlier-robust FSAD-NET'.
During the first training stage of the 'outlier-robust FSAD-NET', we train the DenseNet using the whole training data from Tab.~\ref{tab:polyp_detection_dataset}, and then we train the FSAD-NET using the training data not rejected from the first stage.
Figure~\ref{fig:results_outlier} shows the mean AUC results using the whole testing set from Tab.~\ref{tab:polyp_detection_dataset} (that contains the the Water\&Feces frames) as a function of the number of abnormal (i.e., containing polyp) training images.
This figure compares the outlier-robust FSAD-NET with the original FSAD-NET trained with the whole training set in Tab.~\ref{tab:polyp_detection_dataset}. 
The results indicate that the outlier-robust FSAD-NET achieves better performance for all cases, suggesting that the detector of feces and water jet sprays can effectively exclude these frames.  With 60 abnormal images in the training set, the outlier-robust FSAD-NET achieves an AUC of 97.46\%, which is close to human-level performance from medical practitioners. 

\begin{table}[t!]
\centering
\caption{Results on the data set that does not include Water\&Feces frames, where we compare our proposed FSAD-NET with several zero-shot anomaly detection methods and state-of-the-art and few-shot anomaly detection methods.}
\label{tab:result_polyp_detection}
\begin{tabular}{c|c|c}
                           & Methods                                    & AUC    \\ \hline
 & DAE~\cite{masci2011stacked}                & 0.6384 \\
                           & VAE~\cite{doersch2016tutorial}             & 0.6528 \\
                           & OC-GAN~\cite{perera2019ocgan}              & 0.6137 \\
     Zero-Shot                      & f-AnoGAN(ziz)~\cite{schlegl2019f}          & 0.6629 \\
                           & f-AnoGAN(izi)~\cite{schlegl2019f}          & 0.6831 \\
                           & f-AnoGAN(izif)~\cite{schlegl2019f}         & 0.6997 \\
                           & ADGAN~\cite{liu2020photoshopping}          & \textbf{0.7391} \\ \hline
 & Densenet121~\cite{huang2017densely}   & 0.8231 \\
                           & cross-entropy        & 0.7115 \\
                           & Focal loss~\cite{lin2018focal}            & 0.7235 \\
       Few-Shot                    & without RL            & 0.6011 \\
                           & Learning to Reweight~\cite{ren2018learning}  & 0.7862 \\
                           & AE network           & 0.835  \\
                           & FSAD-NET~\cite{tian2020few}            & \textbf{0.9033}
\end{tabular}
\end{table}



\subsection{Polyp Localisation and Classification Experiments}

The polyp localisation and classification experiment using the Australian and Japanese data sets described in Sec.~\ref{sec:polyp_localisation_classification_dataset} relied on a 5-fold cross validation experiment, where in each fold, the training set contains images from $60\%$ of the patients, the validation set has images of 20\% of patients and the test set contains images of the remaining 20\% of the patients.
The experiment based on training on the Australian data set and testing on the Japanese data set uses the five models learned from the 5-fold cross validation procedure to classify all images from the Japanese data set.  
Results are shown in terms of the mean and standard deviation of accuracy and AUC for classification, using the 5-fold cross validation experiment.  
For polyp localisation, we present our results on the MICCAI 2015 polyp detection challenge~\cite{bernal2017comparative} (also described in Sec.~\ref{sec:polyp_localisation_classification_dataset}), using the following measures: number of true positives and false positives, mean precision and recall, and F1 and F2 scores.

For the polyp localisation and classification approach in~\eqref{eq:classifier_localiser} presented in Sec.~\ref{sec:localisation_classification_polyps}, we use Resnet50~\cite{he2016deep} as the base model for Retinanet~\cite{lin2018focal}, trained with the focal loss function~\eqref{eq:focal_loss}~\cite{lin2018focal}.  
This base model is pre-trained using ImageNet~\cite{deng2009imagenet} because it shows more accurate results than if we train from scratch~\cite{bar2015chest}.   We also rely on data augmentation to improve generalisation, where we increase six fold the training set size using random transformations (rotation, scaling, shearing, and horizontal flipping).  The  validation set is used to tune the hyper-parameters, where the selected values are: 800 epochs, batch size of 32, Adam optimiser, learning rate of $10^{-4}$, dropout rate of 0.3, $\alpha=0.25$ and $\gamma=2$ in~Sec.~\ref{sec:localisation_classification_polyps}.

\begin{figure}[t]
\begin{center}
\includegraphics[width = 1.0\linewidth]{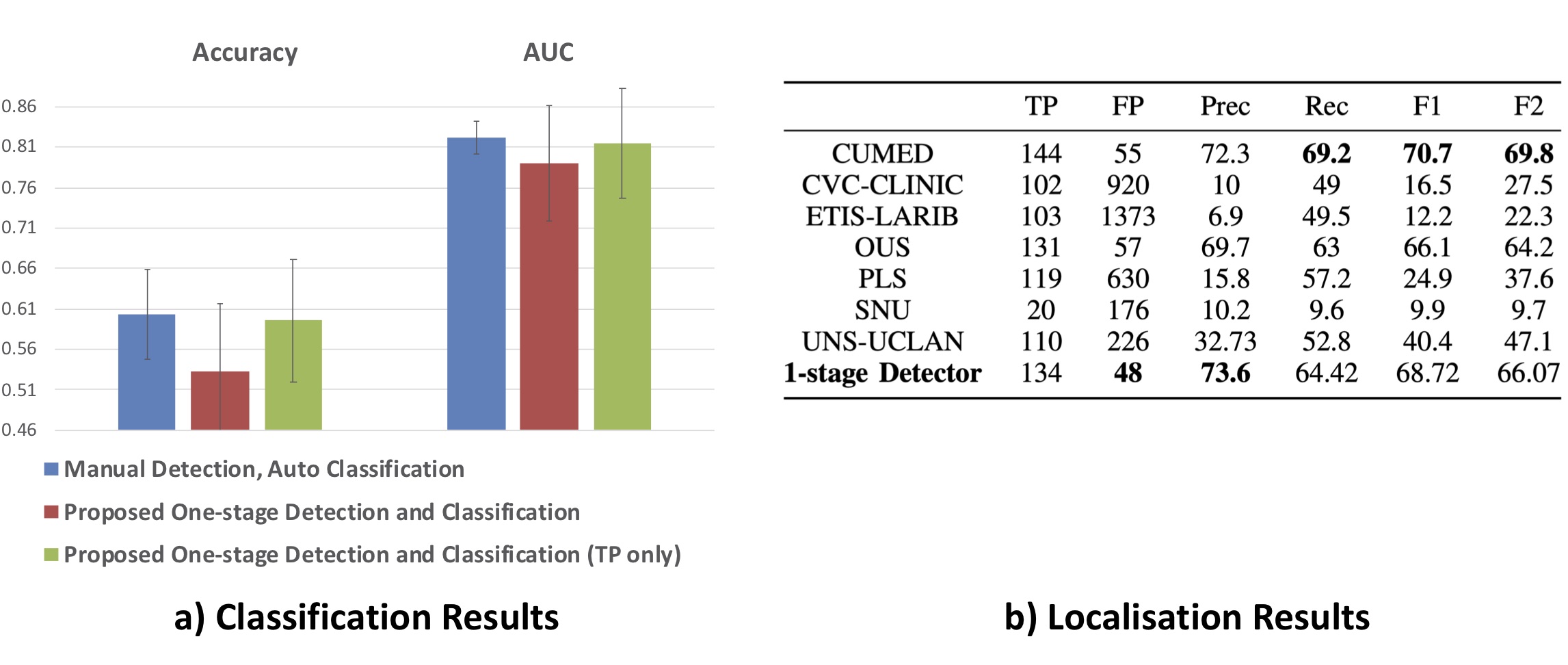}
\end{center}
\caption{The classification results in (a) shows a comparison between our proposed one-stage localisation and classification method and a classification method that uses manual localisation and automated classification of polyps~\cite{pu2018sa1908} using the mean and standard deviation of the classification accuracy and AUC over the 5-fold cross validation experiment.  In this bar plot, we also compute the results from our method considering only the true positive (TP only) polyp localisations to isolate the performance of the classifier.  The localisation results in (b) displays a comparison between our one-stage detector and the state of the art from the MICCAI 2015 Polyp detection Challenge~\cite{bernal2017comparative}.}
\label{fig:results_localisation_classification}
\end{figure}

 The results in Figure~\ref{fig:results_localisation_classification}-(a) shows the classification performance of our proposed model, using as baseline a classifier that relies on manually detected polyps~\cite{pu2018sa1908}.  This baseline is based on a Resnet-50 classifier trained and tested on manually localised polyps, which means that it can only handle correct localisations, so it can be considered as an upper-bound to the performance of our approach. 
If we restrict our method to produce classification results only for true positive localisations (e.g., the bounding boxes have an IoU greater than 0.5 with polyps), 
then the result in Fig.~\ref{fig:results_localisation_classification}-(a) shows that the performance of our method and the baseline are comparable.  On the other hand, when we do not make such restriction, allowing potential false-positive localisations to contaminate the classification process of our approach, then Fig.~\ref{fig:results_localisation_classification}-(a) shows that the baseline is superior.  These results suggest that the classification mistakes made by our approach are mainly due to incorrect localisations.
The polyp localisation results in Figure~\ref{fig:results_localisation_classification}-(b) suggest that although our method is implemented for localising and classifying polyps, it is competitive with the state of the art designed specifically for localising polyps.

%

\subsection{Uncertainty Estimation and Calibration Experiments}

The polyp classification uncertainty and calibration from Sec.~\ref{sec:uncertainty_calibration} uses an ImageNet pre-trained DenseNet-121~\cite{huang2017densely} as backbone.
For the non-Bayesian learning models, the training consists of removing the last 1000-node layer from the pre-trained model and replacing it by a softmax activated five-node layer, representing the five classes of the polyp classification problem.
For the Bayesian learning models, we use concrete dropout~\cite{gal2017concrete}, where the 1000-node layer from the original model is replaced by two fully connected layers: one layer with five nodes activated by a rectified linear unit (ReLU)~\cite{nair2010rectified} and a second layer with ten nodes (first five nodes activated by softmax, representing the classification probability vector, and the next five nodes denoting the aleatoric uncertainty~\cite{kendall2017uncertainties}) -- see Fig.~\ref{fig:classification_localisation_models}-(b).  The parameters of the variational distribution $Q(\theta_P| \psi_Q)$, represented by the mean values of the weights and the dropout probabilities~\eqref{eq:BE} are learned only for these two last layers.

For all training procedures, we use mini-batches of size $32$, $800$ training epochs, initial learning rate of $10^{-3}$, which decays by $0.9$ after every 50 training epochs, and $10 \times$  data augmentation using random translations and scalings.  
The input image size is $224 \times 224 \times 3$ (the original polyp images acquired from colonoscopy videos are transformed to this size by bicubic interpolation).  For the optimisation, we use Adam~\cite{kingma2014adam} with $\beta_1 = 0.9$, $\beta_2 = 0.999$ and $\epsilon = 10^{-8}$.  
For Bayesian inference, the number of samples drawn from $Q(\theta_P| \psi_Q)$ in \eqref{eq:BE_inference} is $N=10$.  
For training the confidence calibration, we retrain the last layer of the model for 100 epochs, using the validation set $\mathcal{V}^{(c)}$ to estimate $s$ in~\eqref{eq:temp_scaling}.
The models tested in this section are labelled as follows: 1) Bayesian learning and inference models in~\eqref{eq:BE} are labelled with \textbf{-Bayes} and non-Bayesian models have no label, and 2) models trained with confidence calibration are labelled as \textbf{+Temp. Scl.} and without calibration as \textbf{+No Scl.}. The combination of these models form a total of four models.

Our classifier that rejects samples using the condition in~\eqref{eq:rejecting_testing} is trained by estimating the hyper-parameters $\tau^*_1(Z)$ and $\tau^*_2(Z)$ using the validation set $\mathcal{V}^{(c)}$, where $Z$  in~\eqref{eq:rejecting_testing} is set as $Z \in \{ 0.5, 0.6, ..., 0.9, 1.0 \}$ (for the training and testing on the Australian data set) and $Z \in \{ 0.7, 0.8, 0.9,1.0 \}$ (for the training on the Australian and testing on the Japanese data set -- the range for this Japanese data set is smaller because of the smaller size of the data set).  For this classifier, the uncertainty is computed from the classification entropy~\eqref{eq:entropy} and confidence is calculated by $P( y | \mathbf{x}, \theta_P^*)$ from~\eqref{eq:basic_model}, $Q( y | \mathbf{x},\psi_Q^*)$ from~\eqref{eq:BE_inference} or the calibrated classifiers from~\eqref{eq:temp_scaling}. The experiments show results where both conditions are applied jointly.

The proposed classifiers in this section are assessed with accuracy and average precision (AP), where accuracy is represented by the proportion of correctly classified samples, and AP is computed by averaging the precision across all recall values between zero and one, and then calculating the mean AP over the five classes.  
The calibration results are calculated with the  expected calibration error (ECE) and maximum calibration error (MCE), where both measures are computed from the reliability diagram, which plots sample classification accuracy as a function of expected accuracy~\cite{guo2017calibration}. These measures are based on the mean (for ECE) or maximum (for MCE) difference between the classification and expected accuracies.

\begin{figure}[t]
\begin{center}
\includegraphics[width=0.49\textwidth]{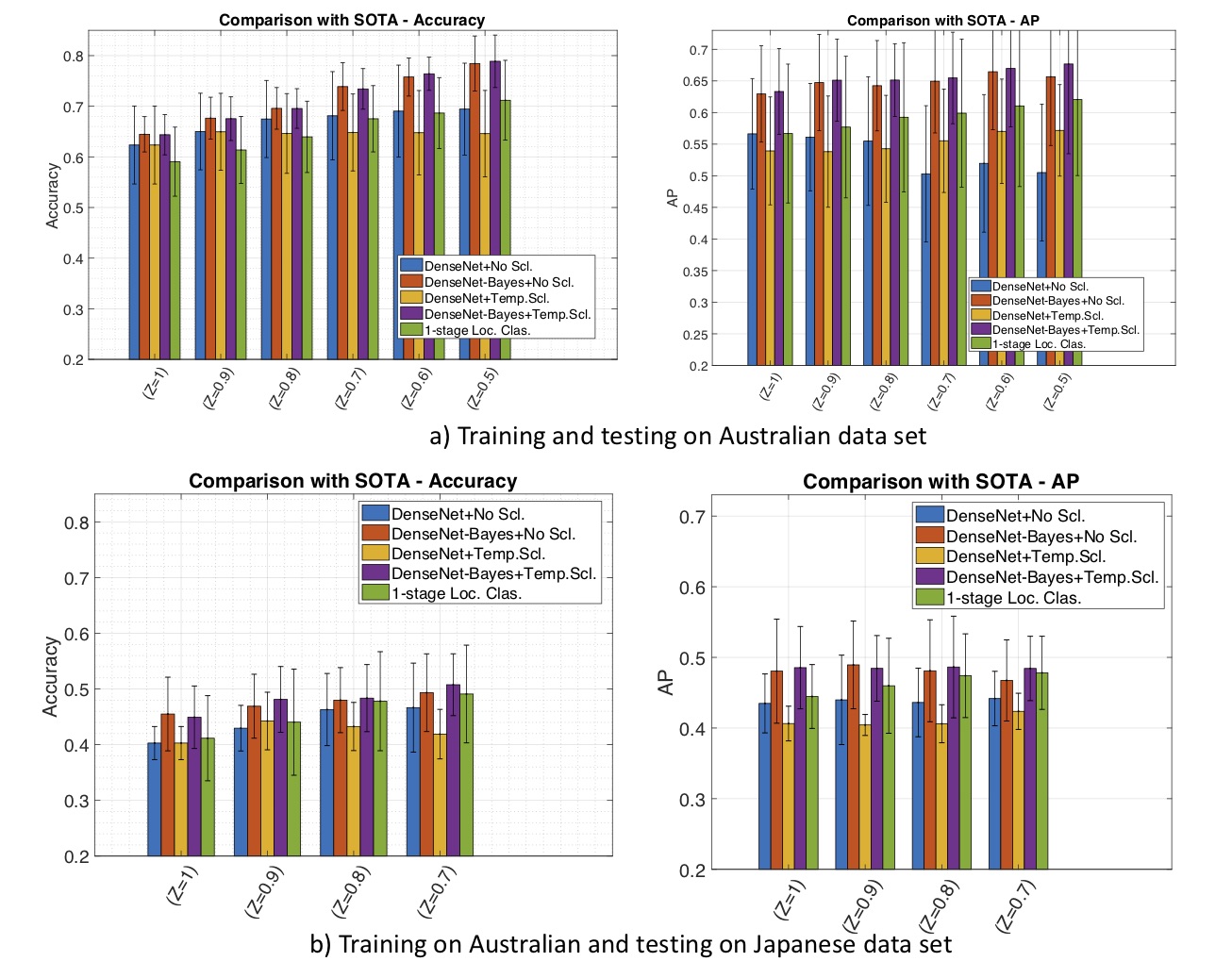} 
\end{center}
\caption{Relying on training and testing from the Australian data set, we show in (a) the mean and standard deviation of the accuracy and mean average precision (AP) results over the 5-fold cross validation experiment for several versions of our proposed method (with different models, calibration and learning algorithm). 
This graph also displays a comparison (in terms of accuracy and AP) between the method
to reject high-uncertainty and low-confidence samples in~\eqref{eq:rejecting_testing} for all models and also for the '1-stage Loc. Clas.' using manually detected polyps.
The same results for the training on Australian set and testing on Japanese set are displayed in (b).}
\label{fig:classification_results_uncertainty}
\end{figure}

Figure~\ref{fig:classification_results_uncertainty}-(a) displays the accuracy and AP using the training and testing sets from the Australian data set, and Fig.~\ref{fig:classification_results_uncertainty}-(b) shows results from the training on Australian and testing on Japanese data set -- all bar plots are grouped as function of the rejection percentage $Z$ from~\eqref{eq:rejecting_testing}.  
We also compare with our polyp localisation and classifier from~\eqref{eq:classifier_localiser}, labelled as '1-stage Loc. Clas.', where the polyp localisation is manually provided and samples are rejected based solely on the first condition in Eq.~\ref{eq:rejecting_testing} (i.e., classification probability).

The results in Fig.~\ref{fig:classification_results_uncertainty} show that the Bayesian DenseNet (with and without calibration) produces the most accurate results.
It is worth noticing the discrepancy in accuracy improvement (as a function of $Z$), compared with AP improvement. 
This can be explained from the imbalanced training set introduced in Sec.~\ref{sec:polyp_localisation_classification_dataset}, with around $40\%$ of the training samples belonging to class $II$ and $20\%$ to class $IIo$.  This biases the classification probability towards these two classes, which explains the better improvement for accuracy, in particular for the Australian set experiment.  
Also from Fig.~\ref{fig:classification_results_uncertainty}, it is clear that accuracy and AP results reduce significantly for the experiment based on training on Australian and testing on Japanese data set, indicating that further studies are necessary to improve the generalisation of the proposed method to new data domains.  
The Bayesian models that reject high-uncertainty and low-confidence samples, trained and tested on the Australian data set, show the best results, with DenseNet-Bayes+Temp.Scl. showing the best overall improvement.  
The comparison with '1-stage Loc. Clas.' shows that the DenseNet-Bayes methods without rejecting samples have superior classification accuracy and AP. 
After rejecting samples based on uncertainty and calibrated confidence, DenseNet-Bayes methods reaches around $70\%$ when rejecting around $20\%$ of the testing samples and close to $80\%$ when rejecting $50\%$ of the testing samples.  The AP also improves, reaching around $64\%$  when rejecting $20\%$ of the testing samples and around $68\%$ when rejecting $50\%$ of the testing samples.  
The rejection process for '1-stage Loc. Clas.', based on uncalibrated confidence, also shows improvements, reaching $71\%$ accuracy and $62\%$ AP when rejecting $50\%$ of the testing samples, but this result is not competitive to the results produced by the DenseNet-Bayes methods.
For the experiment with the training on Australian and testing on Japanese data sets, the classification accuracy of the DenseNet-Bayes methods starts at around $45\%$ and reaches $51\%$ when rejecting around $30\%$ of the testing samples.  This compares favourably with '1-stage Loc. Clas.' that has accuracy of $41\%$ with all samples, and reaches $49\%$ when rejecting $30\%$ of the testing samples.  
Regarding AP, results of the DenseNet-Bayes methods are stable at around $48\%$ with the rejection of testing samples, while '1-stage Loc. Clas.' improves from $44\%$ to around $48\%$ when rejecting around $30\%$ of the testing samples, suggesting that this experiment is more challenging for the DenseNet-Bayes methods, particularly at small values for $Z$.

\begin{figure}[t]
\begin{center}
\includegraphics[width=0.49\textwidth]{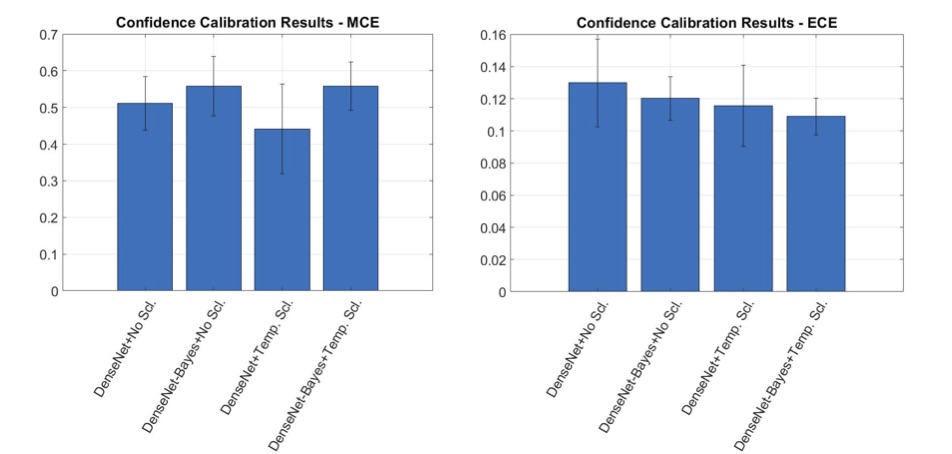} 
\end{center}
\caption{Mean and standard deviation of the MCE (left), and ECE (right) of the methods over the 5-fold cross validation experiment for the training and testing on Australian data set.}
\label{fig:calibration_results_uncertainty}
\end{figure}

The  ECE and MCE results are displayed in Fig.~\ref{fig:calibration_results_uncertainty}.  The calibrated methods tend to show smaller ECE and MCE, but the differences are more noticeable for the non-Bayesian methods. This can be explained by the fact that Bayesian methods tend to produce relatively calibrated results even without an explicit confidence calibration training.

\begin{table}[]
\centering
\caption{The inference time of our proposed methods. }
\resizebox{0.66\linewidth}{!}{%
\begin{tabular}{@{}cc@{}}
\toprule
Method                             & Inference Time   \\ \midrule
FSAD-Net                           & 0.016s per image \\
Outlier-robust FSAD-Net           & 0.032s per image \\ 
1-stage localisation and detection & 0.067s per image \\
Calibration and Uncertainty        & 2.170s per image \\
\bottomrule
\end{tabular}%
}
\label{tab:inference-time}
\end{table}

\subsection{System Running Time}

We measure the run time of all steps of the system for inference in one image. Table~\ref{tab:inference-time} shows the running time of the FSAD-Net, Outlier-robust FSAD-Net, 1-stage localisation and detection, and Calibration and Uncertainty. The results indicate that our system is applicable in real-time colonoscopy procedure, except for the calibration and uncertainty estimator.

\section{Conclusion}

In this paper, we showed the results of several stages of a system that can detect and classify polyps from a colonoscopy video.
The system first rejects frames containing water and feces that can be considered as distractors.
Then, the detection of frames containing polyps is based on 
the few-shot anomaly detector FSAD-NET.
The FSAD-NET comprises an encoder trained to maximise the mutual information between normal images and respective embeddings and a score inference network that classifies between normal and abnormal colonoscopy frames. 
Results showed that FSAD-NET achieved state-of-the-art few-shot anomaly detection performance on our colonoscopy data set, compared to previously proposed anomaly detection methods and imbalanced learning methods. 
Next, the one-stage polyp localisation and classification method proposed is shown to be an effective approach, and when it is compared with a classification method based on manually detected polyps~\cite{pu2018sa1908}, we noted that the small accuracy gap was due to the mis-detected polyps. 
The results for the polyp classification  method that relies on confidence calibration and uncertainty estimation showed that: 1) confidence calibration reduced calibration errors; and 2) rejecting test samples based on high classification uncertainty and low classification confidence improved classification accuracy and average precision for Bayesian methods.  
These results can motivate the development of better interpretable polyp classification methods that outputs not only a classification, but also confidence and uncertainty results.

In the future, we plan to improve polyp localisation accuracy since it is shown to be the major factor causing a gap in classification accuracy with respect to the method that relies on manual polyp detection. Moreover, the polyp classification accuracy also needs to be improved particularly regarding the classes that are underrepresented in the training set.  These issues can be solved with larger annotated data sets.  Another solution for these problems is based on the use of temporal information from the colonoscopy video, where we seek stable polyp detection, localisation and classification over time. 
Another issue that we plan to address is the slow running time of the confidence calibration and uncertainty estimator.
Finally, the generalisation of the methods to new domains also needs to be explored further, and we plan to address that with domain adaptation~\cite{ganin2015unsupervised} and generalisation methods~\cite{li2017learning}.

\bibliographystyle{IEEEtran}
\bibliography{main}

\end{document}